\documentclass[a4paper, 10pt, conference]{ieeeconf}      
\IEEEoverridecommandlockouts
\usepackage{mathptmx} 
\usepackage{times} 
\usepackage{amsmath} 
\usepackage{amssymb}  
\usepackage{cite}
\usepackage{subfig}
\usepackage{tabularx}
\usepackage{graphicx}
\usepackage{color}

\newcommand{\AddRev}[1]{\textcolor{black}{#1}}

\title{Path Following Control System of Line-of-Sight Guidance for Robotic Dolphin with Multi-Link Mechanism in Underwater Simulator
}

\author{Takumi Asada$^{1}$, Takao Oki$^{2}$,  Hideo Furuhashi$^{2}$, Kenta Tabata$^{1}$, Renato Miyagusuku$^{1}$, and Koichi Ozaki$^{1}$ %
\thanks{$^{1}$Utsunomiya University, 7-1-2 Yoto, Utsunomiya, Tochigi, 321-8585, Japan, {\tt\small tasada381@gmail.com}}%
\thanks{$^{2}$Aichi Institute of Technology, 1247 Yachigusa, Yakusa, Toyota, Aichi, Japan, {\tt\small furuhashi@aitech.ac.jp}}%
}

\begin{document}

\maketitle
\thispagestyle{empty}
\pagestyle{empty}

\begin{abstract}
Biomimetic autonomous underwater vehicle (BAUV) with multi-link mechanism is widely used in aquatic life observation and environmental surveys due to its low power consumption and high maneuverability. An environmental survey requires a path following system that automatically follows specific points. However, the path following system of BAUV is limited, and its evaluation with multi-link mechanism robots has not yet been clarified. The path following system in BAUV requires prior simulation because the model differs depending on the type of biomimetics. In this study, we propose a path following system for BAUVs with a multi-link mechanism and evaluation in underwater simulation. In this result, it was possible to design a path following system suitable for BAUV, determine parameters using a simulator, and evaluate control methods.
\end{abstract}

\section{INTRODUCTION}
Autonomous Underwater Vehicles (AUVs) are widely used in various marine industries, such as deep sea exploration, underwater infrastructure inspection, and environment sampling \cite{malliosAutonomousExplorationConfined2016, kunzDeepSeaUnderwater2008, kondoNavigationAUVInvestigation2004, cruzMARESAUVModular2008}. Traditional AUV propulsion systems are used for thruster propulsion, and are capable of high propulsive force and control through thrust allocation \cite{fossenSurveyControlAllocation2006}. BAUVs are attracting attention for the low energy consumption, high maneuverability, and ecological safety that are difficult in traditional AUV \cite{katzschmann2018exploration, songDevelopmentBiomimeticUnderwater2023a, pereiraRayaBioInspiredAUV2025}. Through imitation of aquatic life, BAUV can be utilized for various environmental surveys, inspections, and other tasks. These tasks require technologies such as path following, navigation, and target tracking. In particular, path following systems are utilized in biomimetic robots using various methods. The coverage path planning (CPP) strategy for reducing energy consumption has demonstrated practical applicability in aquaculture management \cite{liuDevelopmentApplicationCoverage}. The optimal control method of learning model predictive control (MPC) has been shown to effectively complete path planning and path following control tasks for robotic fish \cite{muPathPlanningMultiple2022}. The line of sight (LOS) based method has been proposed for controlling the direction of a robot along a desired path \cite{kelasidiIntegralLineofSightGuidance2017, baiWaypointbasedPathFollowing2025}. These path following systems are used in biomimetic robots, but there are no verifications for robots with multi-link mechanisms. This structures require many parameters, such as forces and moments acting on each joint, which complicates simulation and control algorithms. Therefore, it is important to consider a path following system suitable for BAUV with a multi-link structure for underwater exploration.

Furthermore, BAUV has many joint structures and different types. These are classified into body and/or caudal fin (BCF) type and median and/or pectoral fin (MPF) type, each characterized by high speed and high maneuverability. The propulsion methods are varied, so prior verification through simulation is essential. There are various simulations for underwater robots, and Webots simulation are widely used in the field of biomimetics. Previous studies  have simulated various robots, from BAUVs to soft robots, in water environments \cite{krishnamurthy2009multi, du2021underwater, zhangDynamicTargetTracking2024a}. However, there have been no previous studies utilizing simulations of biomimetic robots with multi-link mechanism in an underwater environment. Furthermore, evaluation of the path following system of BAUV using an underwater simulator has not yet been established.

To overcome these issues, we propose a path following system for BAUV in an underwater simulators. Path following system is applied by combining mapping function and LOS guidance system for BAUV with multi-link structure. The mapping function is used to transfer input from LOS to output from a multi-link structure. These systems are simulated using robotic dolphins with multi-link mechanism. Dolphins have been known for their high swimming and turning capabilities \cite{fish1999review}. Various movements are achieved by using the body's flexibility, flukes, and flippers. Evaluating path following systems using aquatic life capable of such various movements will enable expansion of the system for all types of BAUV. This study is expected for further applications, such as adaptation of different underwater robots and verification of algorithms.

\section{Modeling of the BAUV}
\subsection{Hardware Modeling}
The model and specifications of the robotic dolphin with a multi-link mechanism used in the simulation are shown in Fig. 1 and Table I. The dimensions of the robot are 0.758 m in length, 0.132 m in width, and 0.136 m in height. The robot has seven joints, which are located from the head to the fluke in the order of yaw, pitch, yaw, yaw, pitch, pitch, and pitch axes, respectively. The total weight of the model is 6 kg, and the viscous torque and damping parameters have been adjusted using actual machine. \AddRev{Identification methods and verification processes are determined by previous studies \cite{asada2025performance}.}

\begin{figure}[t] 
  \centering
  \resizebox*{8cm}{!}{\includegraphics{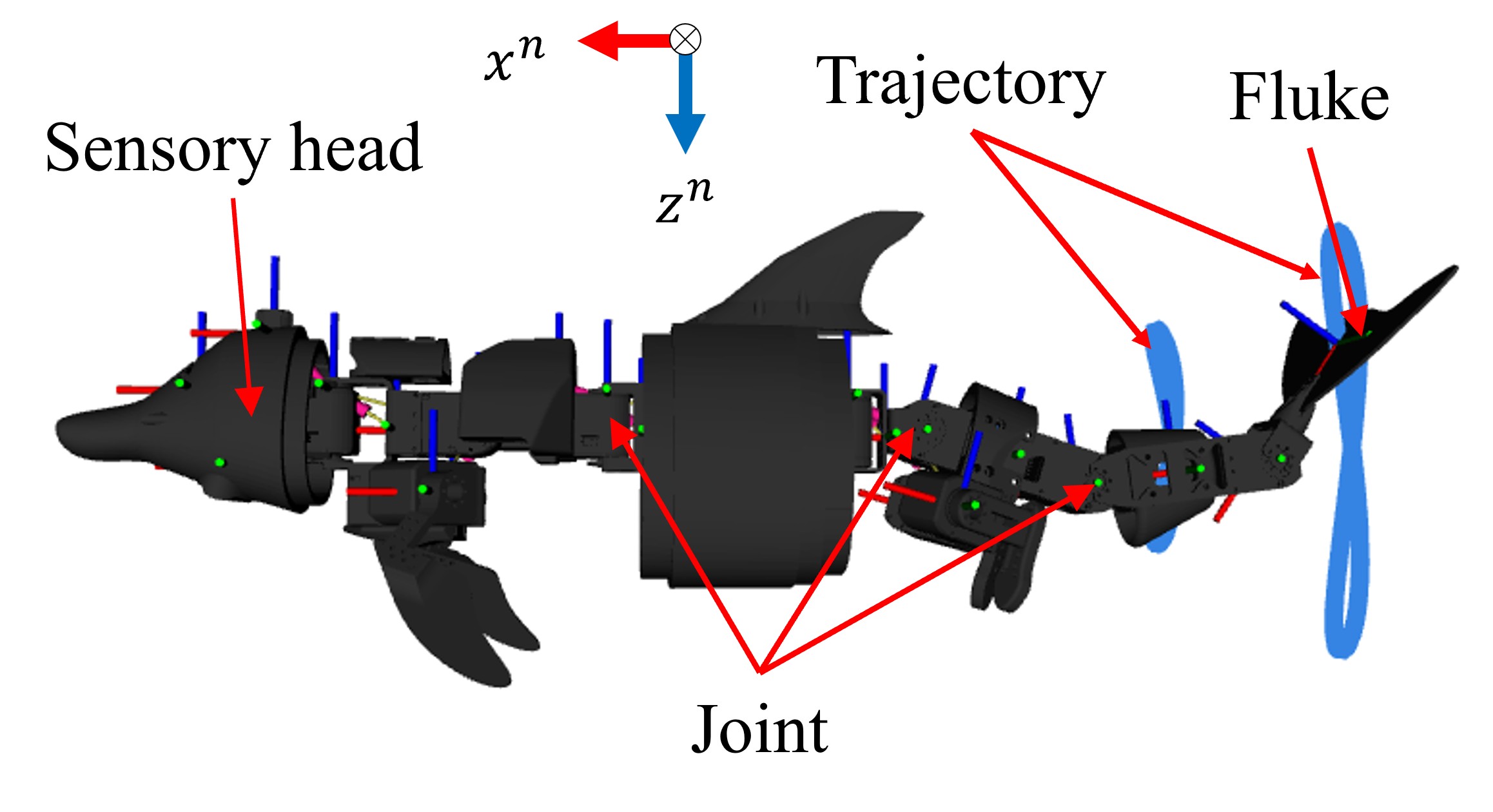}}\hspace{5pt}
  \caption{Model of the robotic dolphin.} \label{fig:Fig1}
\end{figure}
\begin{table}[t]
	\caption{Techinical specifications of the dolphin robot.}
	\label{table:tbl1}
	\begin{center}
	\begin{tabularx}{\linewidth}{ll}
	\hline
	Items & Characteristics \\
	\hline
		Size (L $\times$ W $\times$ H) & 0.758 m $\times$ 0.132 m $\times$ 0.136 m\\
		Total Mass & 6.0 kg \\ 
    Joints & body $\times$  7, fliiper $\times$ 2 \\ 
    Frequency & 0.6 Hz \\
    Viscouse torque & 3 \\
    Damping & 0.5 \\
	\hline
	\end{tabularx}
	\end{center}
\end{table}

\subsection{Software Modeling}
The underwater simulation environment for BAUV is constructed using Webots. The entire system is developed using ROS (Robot Operating System) 2 and communicated with the simulation via the Webots ROS 2 interface. The simulation is constructed from a world model of the underwater environment and a robot driver setting the robot parameters. The robot driver defines damping, drag, and buoyancy parameters for each link. The software diagram for simulation is shown in Fig. 2. \AddRev{The parameters for water density, viscous torque, linear damping, and angular damping are set to
1000 kg/m³, 3, 0.5, and 0.5, respectively.} Each control algorithm is designed from three components: LOS Guidance system, a central pattern generator (CPG) controller, and Vehicle dynamics. The LOS Guidance system sets the target heading angle for a predefined path. The set heading angle is converted by the mapping function, and the joint angles of the robot are generated by the CPG controller. The position of the robot is calculated by solving vehicle dynamics based on the force applied to the robot by the joint angle movement.
\begin{figure}[t] 
  \centering
  \resizebox*{8cm}{!}{\includegraphics{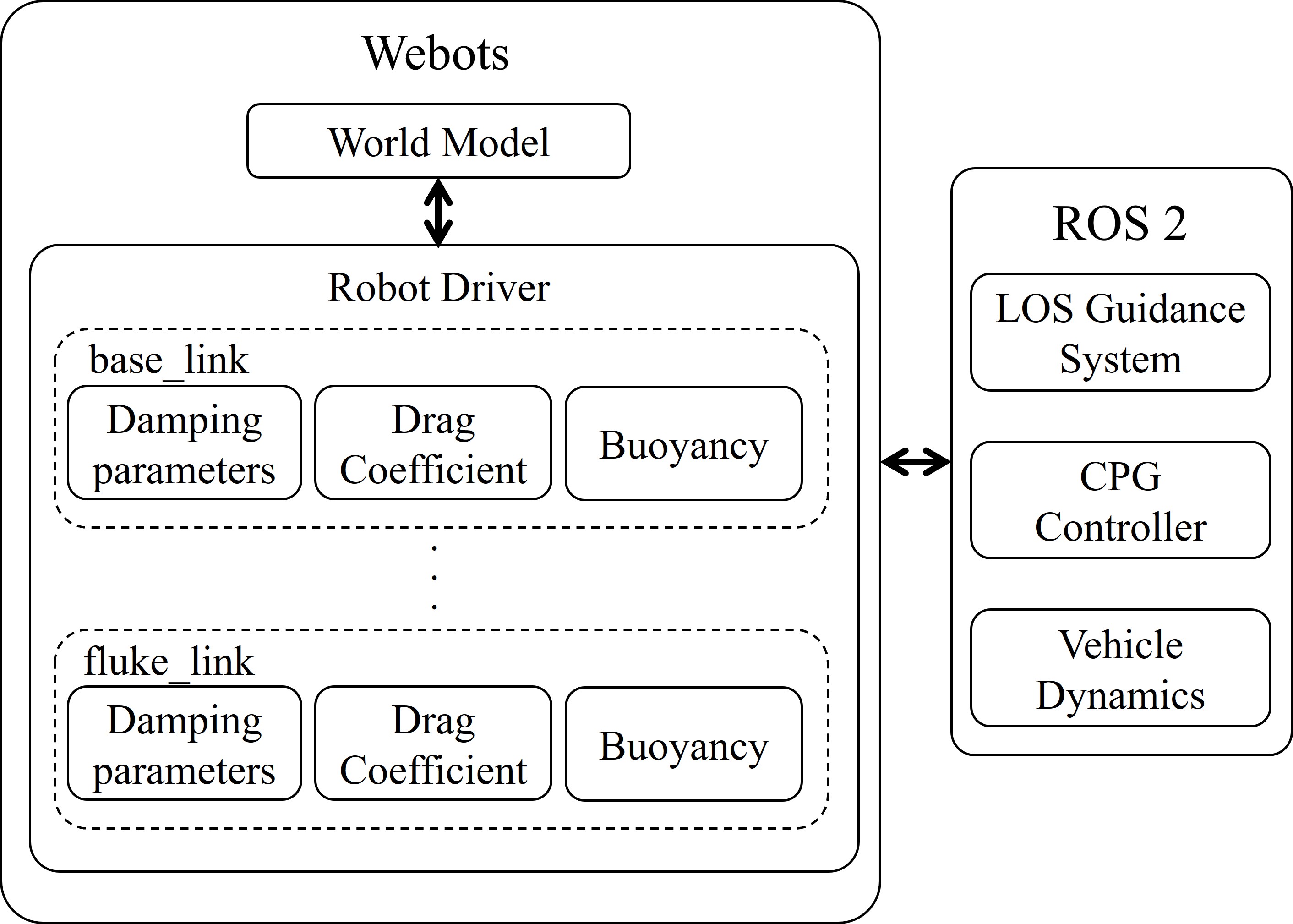}}\hspace{5pt}
  \caption{Software architecture of the robotic dolphin.} \label{fig:Fig2}
\end{figure}

\section{Control Design}
\subsection{Vehicles Dynamics}
In this study, a three-degree-of-freedom (3-DOF) system (surge, sway, yaw) based on the Fossen model \cite{Fossen2021} was used as the BAUV equation of motion. The pitching and rolling motions are ignored because the motion is assumed to be on the water surface. These movements are expressed by Eq. (1) and Eq. (2) as follows:
\begin{equation}\label{eq:eq1}
  \dot{\mathbf{\eta}} = \mathbf{J}(\mathbf{\eta}) \mathbf{\nu}
\end{equation}
\begin{equation}\label{eq:eq2}
  \mathbf{M}\mathbf{\dot{\nu}} + \mathbf{C}(\mathbf{\nu})\mathbf{\nu} + \mathbf{D}(\mathbf{\nu})\mathbf{\nu} = \mathbf{\tau}
\end{equation}

where, the earth-fixed frame $\eta$ is the position and rotation vector expressed by $[x^n, y^n, \psi]^T$. The velocity vector of the body fixed frame $\nu$ is expressed by $[u, v, r]^T$. The transformation matrix $\mathbf{J}(\mathbf{\eta})$, mass matrix $ \mathbf{M}$, Coriolis matrix $ \mathbf{C}$, and drag matrix $ \mathbf{D}$ are given by Eqs. (3)-(5), respectively.
\begin{equation}\label{eq:eq3}
  \mathbf{J}(\mathbf{\eta}) = 
    \begin{bmatrix}
      cos\psi  & - sin\psi & 0 \\
      sin\psi  & cos\psi & 0 \\
      0 & 0 & 1
    \end{bmatrix}
\end{equation}
\begin{equation}\label{eq:eq4}
  \mathbf{M} =
\begin{bmatrix}
  m - X_{\dot{u}} & 0 & 0 \\
  0 & m - Y_{\dot{v}} & 0 \\
  0 & 0 & I_{zz} - N_{\dot{r}} \\
\end{bmatrix}
\end{equation}
\begin{equation}\label{eq:eq5}
  \mathbf{D(\nu)} = -
\begin{bmatrix}
    X_{u} & 0 & 0 \\
    0 & Y_{v} & 0 \\
    0 & 0 & N_{r} \\
\end{bmatrix}
\end{equation}

\AddRev{The Coriolis term can be ignored for the entire BAUV system because its mass and velocity are small.} The parameter $I_{zz}$ indicates the moment of inertia, and $X_u$, $Y_v$, $N_r$, $X_{\dot{u}}$, $Y_{\dot{v}}$, $N_{\dot{r}}$ are the added mass and its derivatives, defined by Lamb's k-factor definition \cite{lamb1924hydrodynamics}. Dolphins perform lift-type propulsion, therefore the force applied to the robot is calculated based on the velocity and angular velocity of each joint. The force $F_{b}$ and moment $M_{b}$ applied to the robot body frame are determined by summing the forces acting on each joint, as shown in Eq. (6):

\begin{equation}\label{eq:eq17}
  \begin{aligned}
      \mathbf{\tau} &= \begin{bmatrix}
        X \\
        Y \\
        N
      \end{bmatrix} = \begin{bmatrix}
        \sum_{i=1}^{7} (F_{b:x,i}) \\
        \sum_{i=1}^{7} (F_{b:y,i}) \\
        \sum_{i=1}^{7} (M_{b:z,i}) 
      \end{bmatrix}
  \end{aligned}
\end{equation}

where $X$, $Y$, and $N$ represent the propulsive force in x- and y-directions, and the moment generated in the z-axis direction, respectively. 

\subsection{Path Following System}

As path following system, the LOS guidance system commonly used in marine craft is utilized \cite{Fossen2021}. A schematic diagram of the path following system using LOS is shown in Fig. 3. Consider a vehicles' position $\mathbf{p}^n = [x^n, y^n]^T$ in the NED frame, following a desired path. In a path-tangential frame, the path-tangential angle is defined by Eq. 7:

\begin{figure}[t] 
  \centering
  \resizebox*{8cm}{!}{\includegraphics{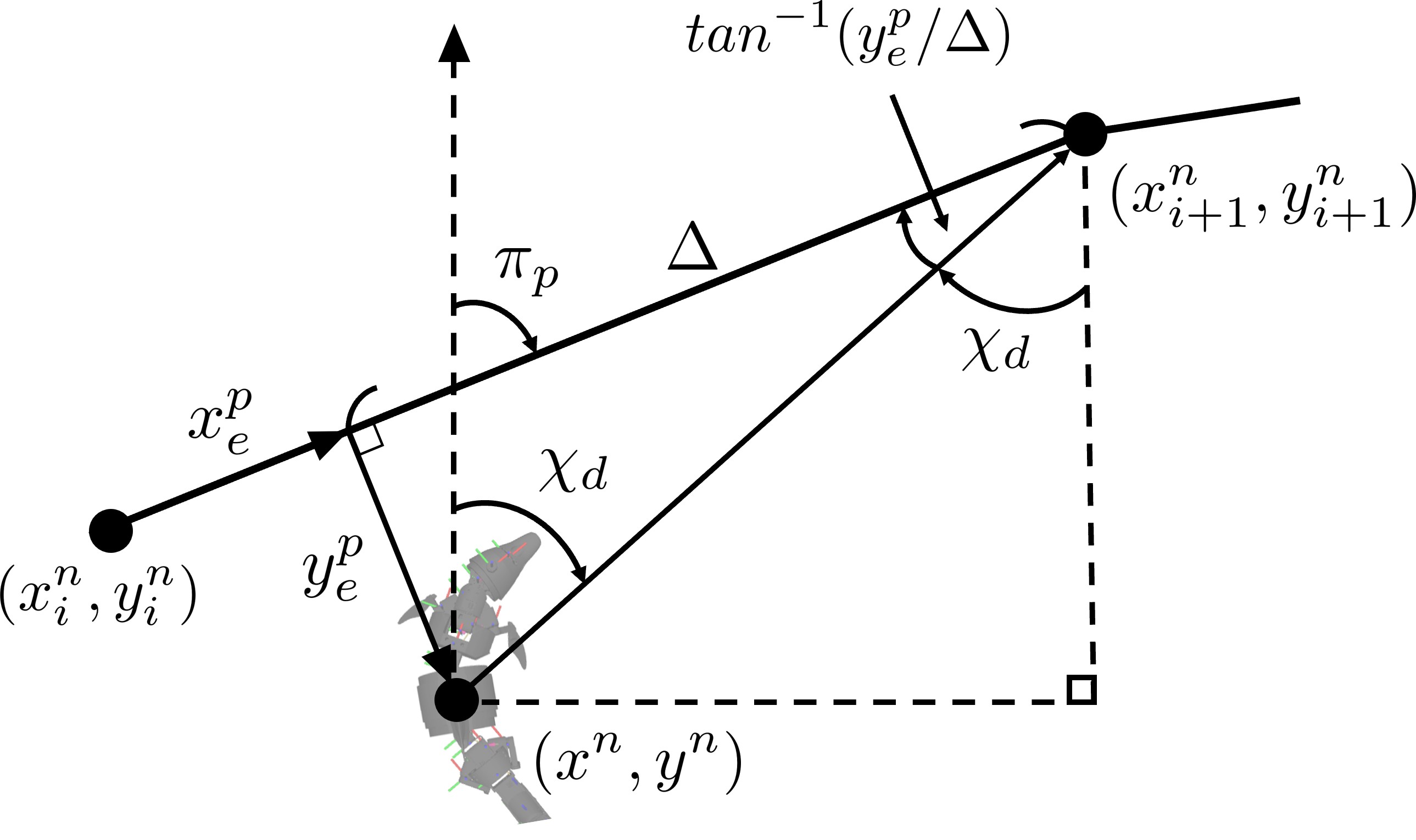}}\hspace{5pt}
  \caption{Schematic diagram of path following system.} \label{fig:Fig3}
\end{figure}
\begin{equation}\label{eq:eq7}
	\pi_p = \arctan(y_{i+1}^n - y_i^n, x_{i+1}^n - x_i^n)
\end{equation}

where $(x_{i}^n, x_{i}^n)$ is the successive waypoint for $i = 1, 2, \cdots N$ in the NED frame. In traditional LOS guidance law, the heading autopilot command $\chi_d$ can be expressed using the path-tangential angle in Eq. 8.

\begin{equation}\label{eq:eq8}
  \chi_d = \pi_p - tan^{-1} (y_e^p/\Delta)
\end{equation}

The parameters $y_e^p$ and $\Delta$ represent the cross-track error expressed in the path-tangential reference frame and look-ahead distance, respectively. The look-ahead distance is $\Delta > 0$ and must be configured appropriately. This is because selecting a closest point will not minimize cross-track error at points with high curvature \cite{atoui2021real}.

The transformation from NED frame to path-tangent frame is given by Eq. (9), and the cross-track error can be obtained by solving Eq. (10):
\begin{equation}\label{eq:eq9}
\begin{aligned}
\mathbf{A} &= \scalebox{0.75}{$
\begin{bmatrix}
cos(\pi_p) & sin(\pi_p) & 0 \\
-sin(\pi_p) & cos(\pi_p) & 1 \\
tan(\pi_p) & -1 & 0
\end{bmatrix},  \quad  \quad  \quad
$} \mathbf{b} = \scalebox{0.7}{$
\begin{bmatrix}
cos(\pi_p) x^n + sin(\pi_p) y^n \\
-sin(\pi_p) x^n + cos(\pi_p) y^n \\
tan(\pi_p) x^n_{i+1} - y^n_{i+1}
\end{bmatrix}
$}
\end{aligned}
\end{equation}
\begin{equation}\label{eq:eq10}
  x = \mathbf{A}^{-1}\mathbf{b}
\end{equation}

where, $x = [x_p^n, y_p^n, y_e^p]^T$ is the origin of the path-tangential reference frame and the cross-track error.

Also, the adaptive LOS (ALOS) guidance law has been used as a model that considers unknown disturbances caused by wind, waves, and ocean currents \cite{fossenAdaptiveLineofSightALOS2023}. The desired heading angle of ALOS considering unknown disturbances is calculated as Eq. (11):
\begin{equation}\label{eq:eq11}
\chi_d = \pi_p - \hat{\beta} - \arctan\left(\frac{y_e^p}{\Delta}\right)
\end{equation}

where $\hat{\beta}$ is the estimated sideslip angle expressed by Eq. (12).
\begin{equation}\label{eq:eq12}
  \dot{\hat{\beta}} = \gamma \frac{\Delta}{\sqrt{\Delta^2 + (y_e^p)^2}} y_e^p
\end{equation}
The parameter of $\gamma > 0$ is the adaptation gain. Using traditional LOS and ALOS, the path following system of BAUV with a multi-link structure is compared and evaluated. The look-ahead distance is important for the heading autopilot command, so it is configured with a values that is based on the total robot length.

\subsection{CPG Controller}
The BAUV locomotion with multi-link mechanism is controlled using a CPG controller. This controller is used in various biomimetic robots for rhythmic motion generation. The robot using the simulation has a multi-link mechanism with a total of seven joints ($J_i: i=1, \cdots, 7)$ on the pitch and yaw axes. Therefore, CPG models are needed that allow parameters to be easily changed. Therefore, it is controlled by the CPG model in Eq. (13), which is commonly used for robots with multi-link mechanisms such as lamprey robots \cite{angelidis2021spiking}.
\begin{equation}\label{eq:eq13}
  \begin{aligned}
    \dot \phi_i &= 2 \pi f_i + \sum_j \omega_{i,j} sin(\phi_j - \phi_i - \Delta \varphi_{ij}) \\
    \ddot{r_i} &= a_i (\frac{a_i}{4} (\bar{R_i} - r_i) - \dot{r_i}) \\
    \ddot{\chi_i} &= b_i (\frac{b_i}{4} (\bar{X_i} - \chi_i) - \dot{\chi_i}) \\
    \theta_i &= \chi_i + r_i sin(\phi_i)
  \end{aligned}
\end{equation}

The parameter of $\phi_i$, $r_i$, $\chi_i$, $\theta_i$ are the phase, amplitude, bias amplitude and output angle of oscillator $i$, respectively. CPG parameters are defined in advance, considering the maximum joint angle, desired amplitude, and offset. 

A mapping function is used to automatically control the robot's movements for the desired heading angle input from LOS. The robotic dolphin is equipped with three yaw axes ($J_1, J_3, and J_4$), and desired command values are given to each joint. To smoothly convert LOS input to azimuth direction signs, a hyperbolic tangent function is used. The conversion from LOS input to desired offset angle of yaw angle is shown in Eq (14).
\begin{equation}\label{eq:eq14}
\begin{aligned}
  \bar{X_i} &= \frac{e^{2k \chi_d} - 1}{e^{2k \chi_d} + 1} X_i
\end{aligned}
\end{equation}

The index $i$ is 1, 3, and 4, and $k$ indicates the gradient of the function. The forward speed is adjusted by changing the desired amplitude of the fluke's tail fin ($J_5, J_6$, and $J_7$) according to the heading angle. The desired amplitude is changed using the Gaussian function shown in Eq. (15) from the input from LOS.
\begin{equation}\label{eq:eq15}
\begin{aligned}
  \bar{R_{i}} &= exp (-\frac 1 2 (\frac {\chi_d-c_{i}} {b_{i}})^2) R_i
\end{aligned}
\end{equation}

Parameters of the gaussian function $b_i $, and $c_i$ are respectively $b_5, b_6$, and $b_7 = 1.0$ and $c_i$ are defined as $c_5 = c_6 = c_7 = 0.0$. During forward movement, the desired amplitudes $R_5, R_6$, and $R_7$ are set to 20, 40, and 60, respectively. The desired offset of yaw angle $\chi_1, \chi_3$, and $\chi_4$ during the turning movement is set to 30.

\subsection{Sinusoidal Path Generation Equations}
A sinusoidal path is used as the waypoint generator for the path following system. The experimental of zig-zag test is used to estimate the uncertain displacement and drag of a robot \cite{klinger2014experimental}. To reproduce the zig-zag test in the propulsion method of robotic dolphins, a sinusoidal trajectory is used. Waypoints are generated from the robot's current position using Eq. (16).
\begin{equation}\label{eq:eq16}
\begin{aligned}
x_i &= \frac{L}{n - 1} \cos(\theta) - A \sin\left(\frac{2\pi N}{n - 1}\right) \sin(\theta) + x_0 \\
y_i &= \frac{L}{n - 1} \sin(\theta) + A \sin\left(\frac{2\pi N}{n - 1}\right) \cos(\theta) + y_0
\end{aligned}
\end{equation}

The parameters $A$, $N$, $L$, $\theta$, and $n$ are the amplitude, sinusoidal period, total path length, heading angle of the base robot, and total number of path points, respectively. The path origin point $(x_0, y_0)$ is used to generate a path from the current position of the robot. The diagram generated by path generation is shown in Fig. 4. In the underwater experimental simulation, parameters of path generation are set to $A$ = 0.5 m, $N$ = 3, and $L$ = 10 m.
\begin{figure}[t] 
  \centering
  \resizebox*{8cm}{!}{\includegraphics{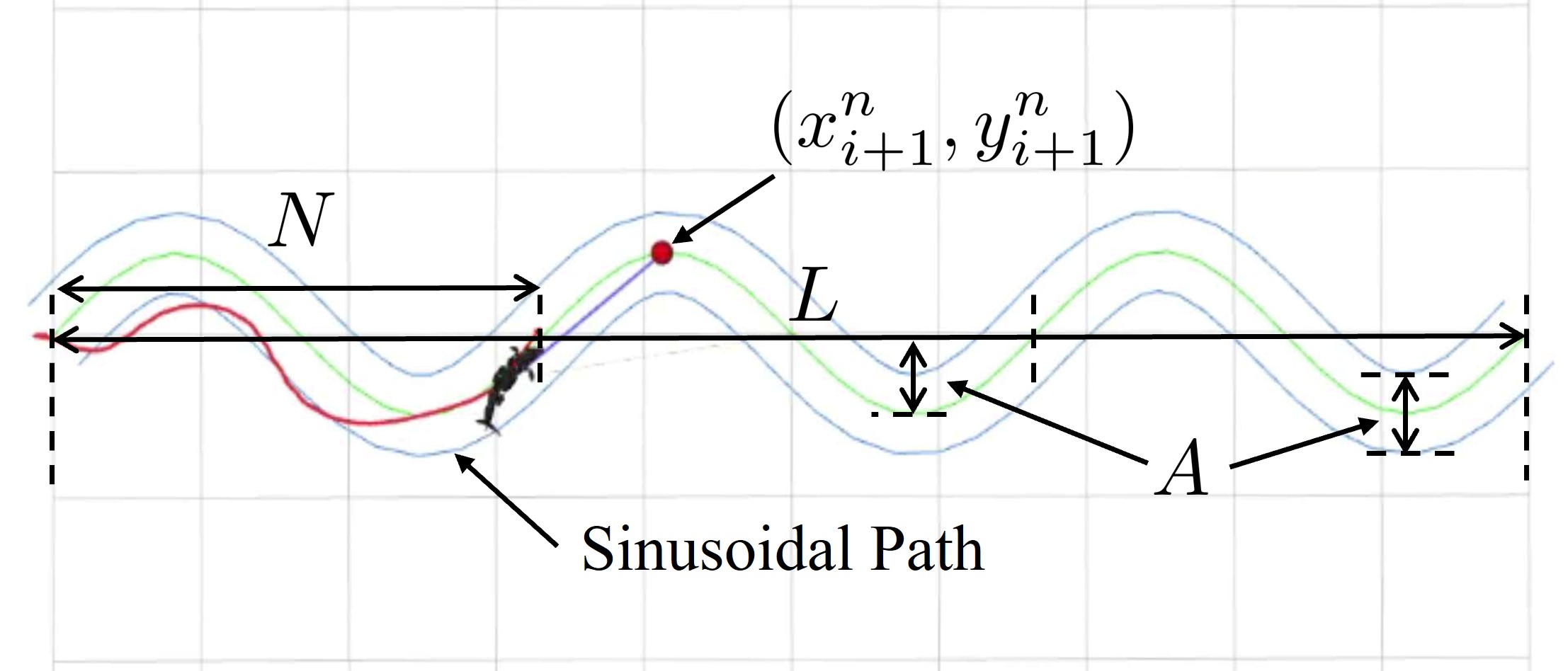}}\hspace{5pt}
  \caption{Diagram of a sinusoidal path.} \label{fig:Fig4}
\end{figure}
From the above, the path following system of BAUV with a multi-link mechanism can be controlled by integrating these systems. The control block diagram of the overall system is shown in Fig. 5. \AddRev{State feedback $x^n$, $y^n$ can be calculated using GPS measurements or by solving Fossen's vehicle dynamics model. The response delay of the actuator and the affect of friction due to waterproofing are considered small and ignorable.}
\begin{figure}[t] 
  \centering
  \resizebox*{8cm}{!}{\includegraphics{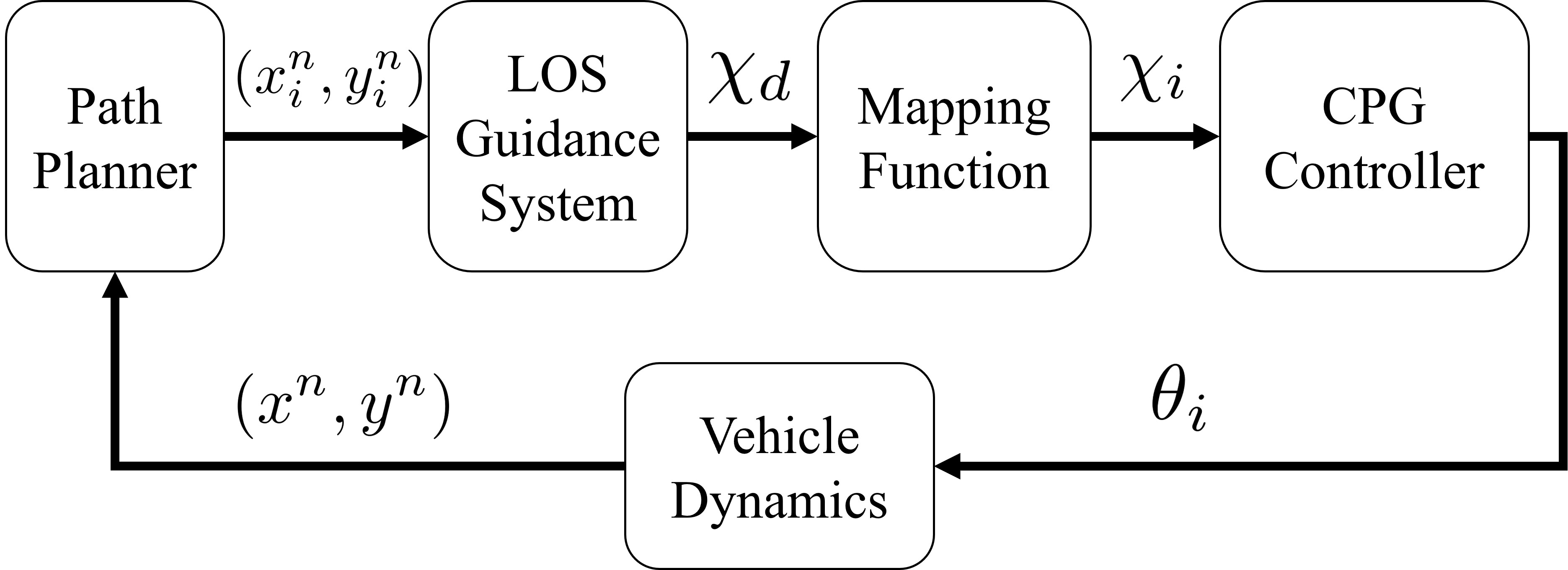}}\hspace{5pt}
  \caption{Control block diagram of overall system.} \label{fig:Fig5}
\end{figure}

\section{Experimental Results in Underwater Simulation}
\subsection{Path Following Experiments}
Experiments on the path following system are conducted in an underwater simulation environment. The simulation environment is shown in Fig. 6. The dimensions of the underwater experimental pool are 15 [m] in length, 2 [m] in width, and 0.45 [m] in depth. The experiment is conducted by varying the amplitude mapping from the desired heading angle and the look-ahead distance. The amplitude is increased or decreased for multiple fluke joints by a mapping function, so it is implemented in two patterns: maximum amplitude or amplitude control. The path following system is evaluated using two methods: traditional LOS and ALOS. As each system parameter, the look-ahead distance $\Delta$ is changed by total length $1.5 L, 1.75 L$, and $2.0L$ for experimentation.
\begin{figure}[t] 
  \centering
  \resizebox*{8cm}{!}{\includegraphics{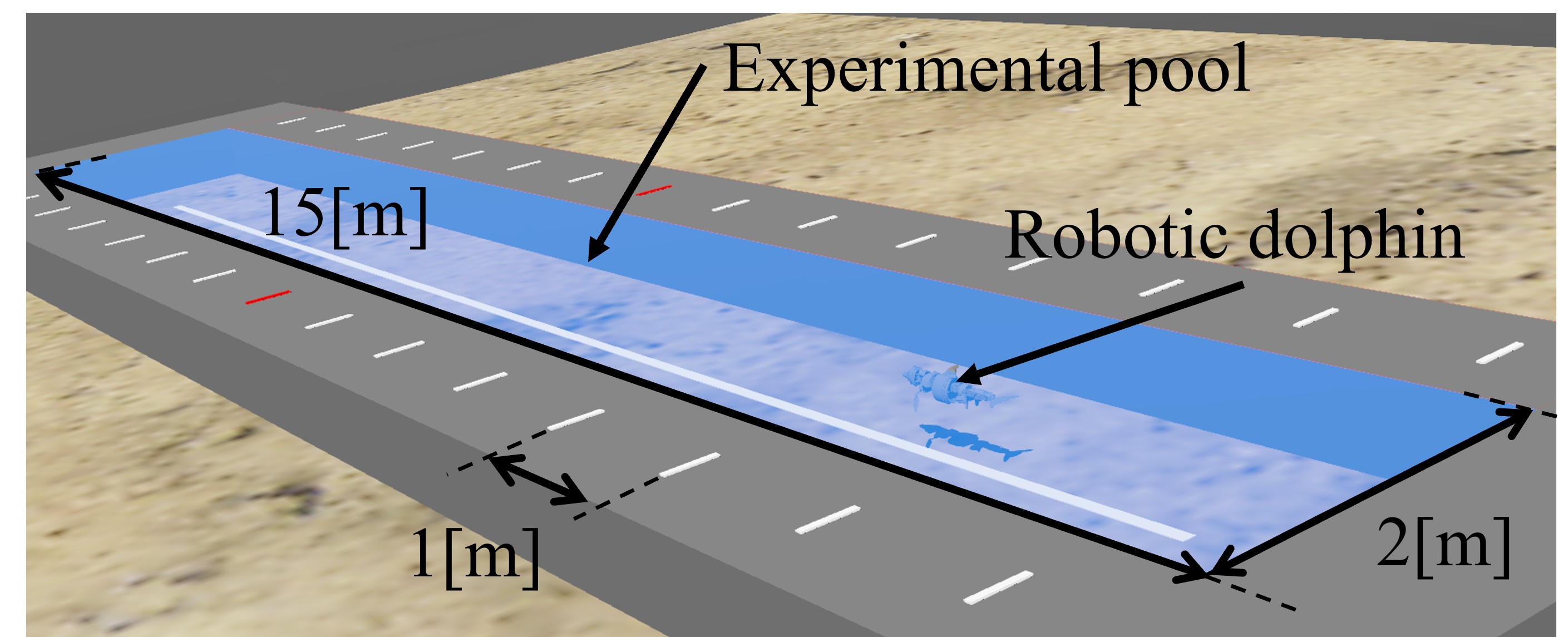}}\hspace{5pt}
  \caption{Underwater simulation environment.} \label{fig:Fig6}
\end{figure}
Representative series of snapshots from the underwater simulation is shown in Fig. 7, and the results are shown in Fig. 8.
\begin{figure}[t] 
  \centering
  \resizebox*{8cm}{!}{\includegraphics{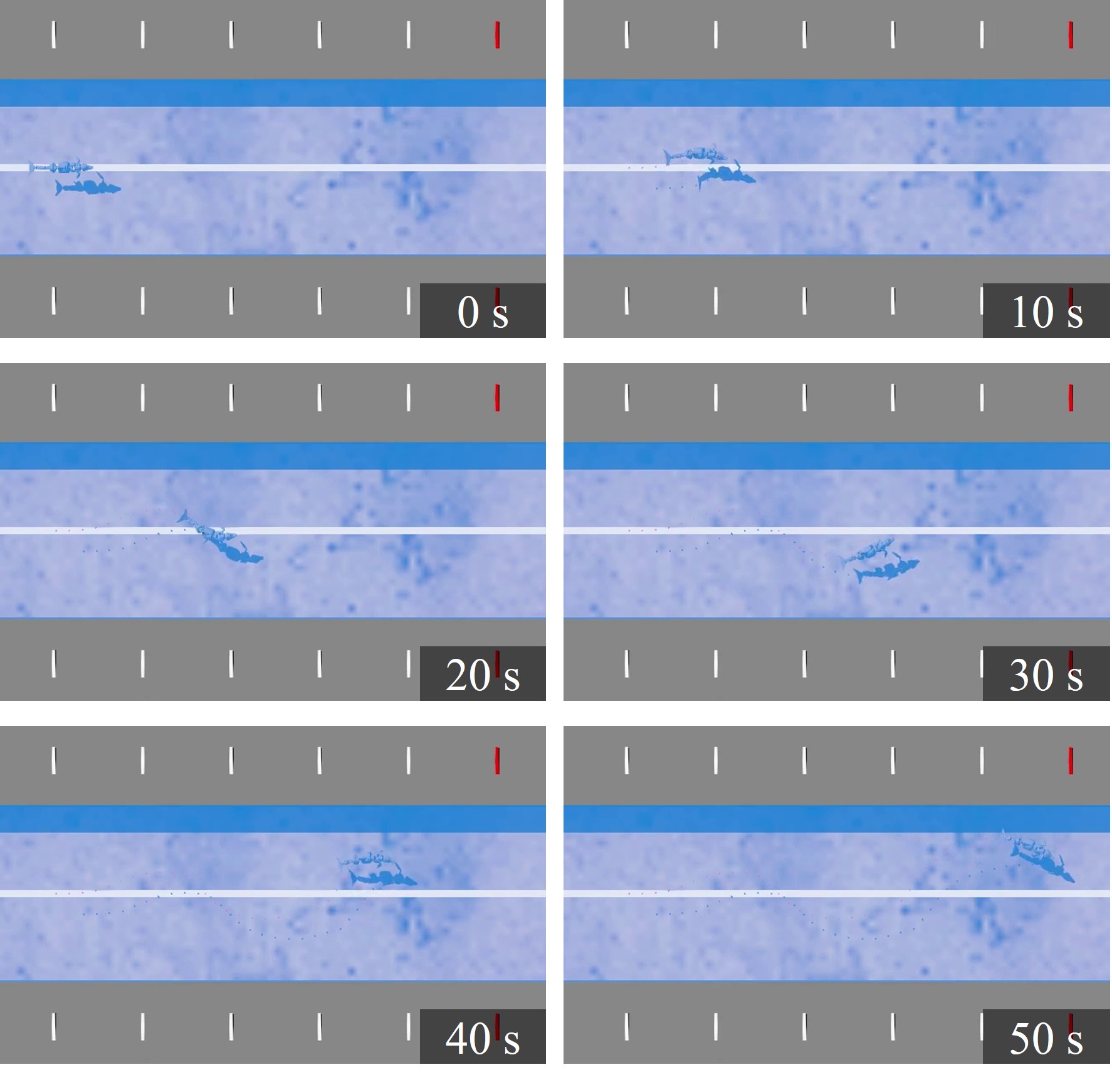}}\hspace{5pt}
  \caption{Snapshot series of the path following.} \label{fig:Fig7}
\end{figure}
\begin{figure}[t] \label{fig:Fig8}
  \centering
  \subfloat[]{
  \resizebox*{8cm}{!}{\includegraphics{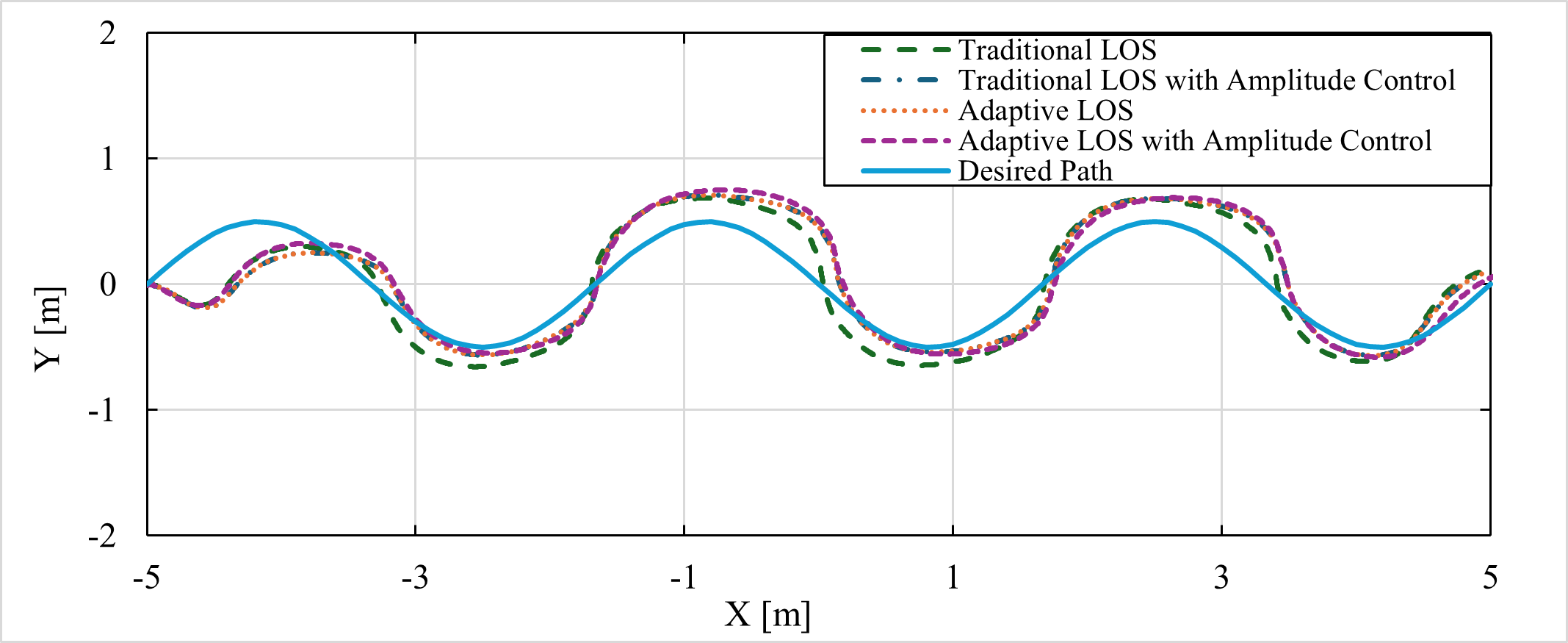}}}\hspace{5pt}
  \subfloat[]{
  \resizebox*{8cm}{!}{\includegraphics{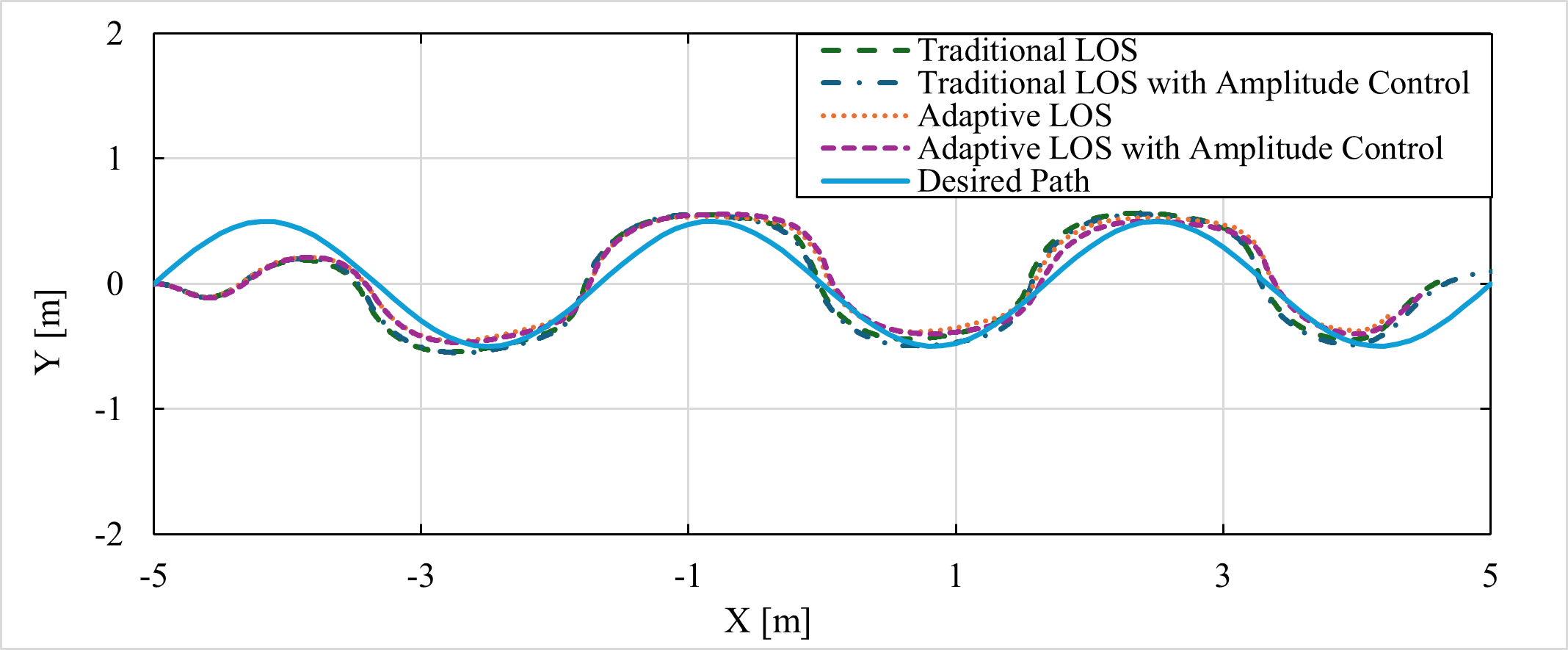}}}\hspace{5pt}
  \subfloat[]{
  \resizebox*{8cm}{!}{\includegraphics{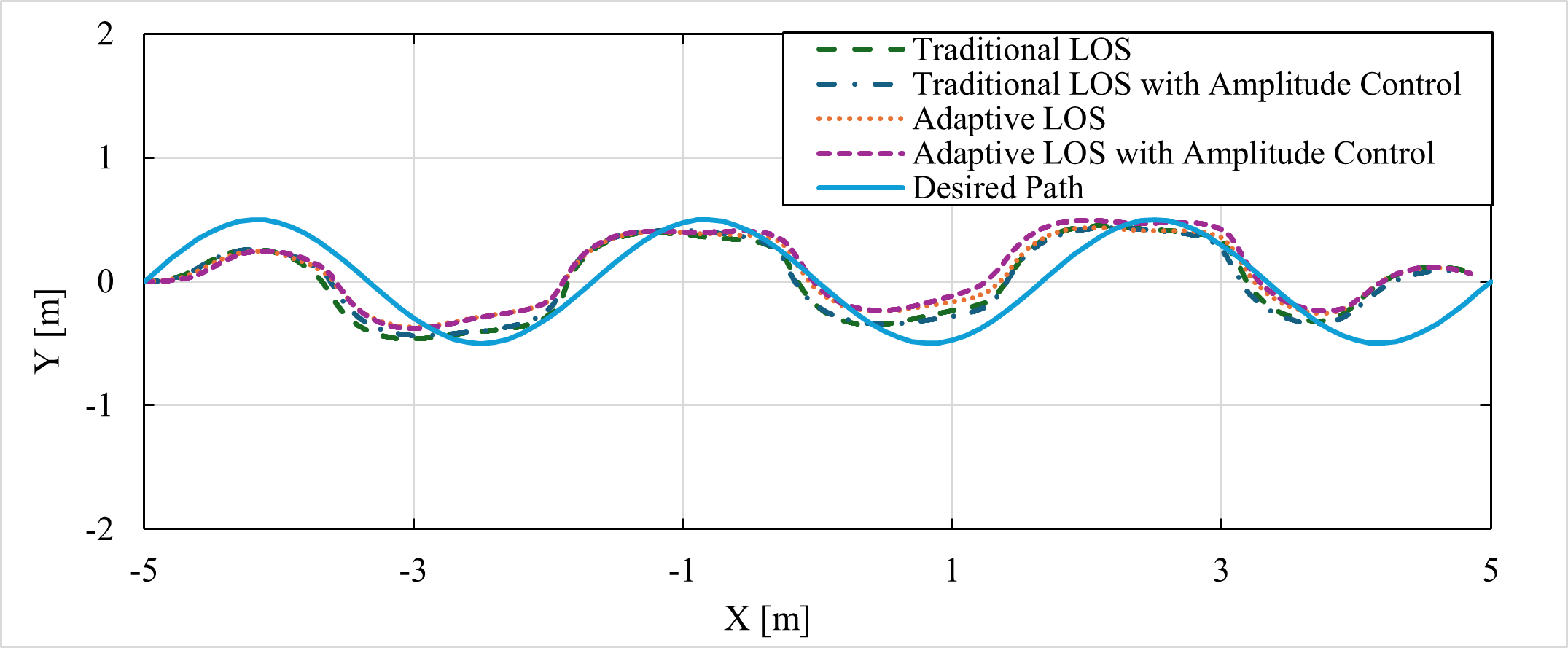}}}\hspace{5pt}
  \caption{Result of path following system (a) $\Delta = 1.5L$, (b) $\Delta = 1.75L$, (c) $\Delta = 2.0L$.}
\end{figure}
\begin{table}[t]
	\caption{Quantitative analysis of the 1.75$L$ tracking error.}
	\label{table:tbl2}
	\begin{center}
	\begin{tabularx}{\linewidth}{lll}
	\hline
	Items & RMSE & MAE \\
	\hline
		Traditional LOS & 0.1351 & 0.1122 \\
    Traditional LOS with Amplitude Control & 0.1181 & 0.0975\\
    Adaptive LOS & 0.1176 & 0.0954\\
    Adaptive LOS with Amplitude Control & 0.0893 & 0.0712\\
	\hline
	\end{tabularx}
	\end{center}
\end{table}
\subsection{Results}
The simulation results of path following in Fig. 8 showed that the desired path was closest with a look-ahead distance of 1.75$L$. The turning radius increases in any pattern of amplitude control and LOS type, given that the look-ahead distance is 1.5 times the total length $L$. A look-ahead distance of 2.0 $L$ is defined, the robot moves along a path that is smaller than the maximum amplitude of the desired path. In 1.5$L$ and 2.0$L$, the straight lines in the path have large errors. This could be considered that the desired heading angle could not be controlled because the look-ahead distance for the waypoint path was too small or large. These results conclude that the relationship between look-ahead distance and the total length of BAUV is necessary for appropriate path following. 

The LOS type evaluation utilizes root mean square error (RMSE) and mean absolute error (MAE) for quantitative analysis. The quantitative analysis results for a look-ahead distance of 1.75 $L$ are shown in Table II. The RMSE and MAE were smaller than those of traditional LOS in the case of a look-ahead distance of 1.75$L$. This also shows that ALOS has smaller errors than traditional LOS at all look-ahead distances. This is because the ALOS system is designed to consider unknown disturbances, resulting in high robustness. In amplitude control using mapping functions, the RSME and MAE are both small when the amplitude is set to the maximum value. This experiment shows that path following is possible with flexible control of the heading angle. It is considered because BAUV has a multi-link mechanism, allowing flexible control by adjusting the yaw axis. The use of underwater simulation to compare and quantitatively analyze algorithms in advance is useful.

\section{CONCLUSIONS}
A path following system for BAUV with a multi-link mechanism was implemented in an underwater simulation environment. The path following system consists of three components and is provided to BAUV by the mapping function. A mapping function was implemented to adapt input from LOS to the multi-link mechanism. Heading angle control using a mapping function enabled control close to the desired path. This function performed satisfactorily as a control output for BAUV with a multi-link mechanism. Also, the appropriate look-ahead distance setting indicates that BAUV follows a path close to the desired path by the mapping function. The path following system was experimented with two types of LOS, amplitude control using a mapping function, and look-ahead distance parameters. As a result, a lookahead distance of 1.75 times the total length of the model is appropriate. It has been clarified that the parameters need to be set appropriately for the total length. This shows the usefulness of adjusting parameters in advance using underwater simulation. Through quantitative analysis using simulation, the ALOS with amplitude control was shown to minimize errors in both RMSE and MAE. Simulation verification in advance of LOS and ALOS amplitude control has been shown to be sufficiently effective in evaluating BAUV performance.
\section{FEATURE WORK}
Through simulation, machine learning and reinforcement learning (RL) methods will be investigated to enable appropriate parameter adjustment according to the relationship between total length and look-ahead distance. \AddRev{Various types of BAUVs are modeled through simulation and trained using RL methods. Based on the movements of actual aquatic life, parameters are identified for the target type using inverse reinforcement learning. This is expected to be utilized as a planning system for all types of BAUVs.} Improvements to the activation function that automatically tunes the CPG parameters of the multi-link mechanism are also expected, based on the desired heading angle.




\bibliographystyle{IEEEtran}  
\bibliography{IEEEabrv,bibtex}

\end{document}